\documentclass{article} 
\usepackage[preprint]{colm2026_conference}

\usepackage{microtype}
\usepackage{hyperref}
\usepackage{url}
\usepackage{booktabs}
\usepackage{amsmath}
\usepackage{graphicx}
\usepackage{multirow}
\usepackage{wrapfig}
\usepackage{adjustbox}

\usepackage{lineno}

\definecolor{darkblue}{rgb}{0, 0, 0.5}
\hypersetup{colorlinks=true, citecolor=darkblue, linkcolor=darkblue, urlcolor=darkblue}

\title{Is There Knowledge Left to Extract? Evidence of Fragility in Medically Fine-Tuned Vision-Language Models}

\author{Oliver McLaughlin$^1$, Daniel Shubin$^2$, Carsten Eickhoff$^3$, Ritambhara Singh$^1$, \\
\textbf{William Rudman}$^{4*}$ \& \textbf{Michal Golovanevsky}$^{1*}$ \\
$^1$Brown University \quad 
$^2$University of Washington \\
$^3$University of Tübingen \quad 
$^4$The University of Texas at Austin \\
\texttt{william.rudman@utexas.edu} \quad \texttt{michal\_golovanevsky@brown.edu}\\
{\small $^*$Equal senior contribution}
}

\begin{document}

\ifcolmsubmission
\linenumbers
\fi

\maketitle

\begin{abstract}
Vision-language models (VLMs) are increasingly adapted through domain-specific fine-tuning, yet it remains unclear whether fine-tuning enables reasoning beyond superficial visual cues, a critical requirement in high-risk settings. This is a particular challenge in the medical domain, where robust reasoning is critical for adoption in real-world clinical settings. In this work, we evaluate four paired open-source VLMs (LLaVA vs.\ LLaVA-Med; Gemma vs.\ MedGemma) across four medical imaging tasks of increasing diagnostic difficulty: brain tumor identification, pneumonia detection, skin cancer classification, and histopathology cancer recognition.
Prior work has shown that VLMs are highly sensitive to input formulation and can rely on heuristic associations learned from data rather than genuine visual understanding, but how these factors interact with domain-specific fine-tuning to affect robustness remains unclear. We find that as task difficulty increases, performance in medically fine-tuned VLMs degrades toward near-random levels, indicating key limitations in clinical reasoning. Medical fine-tuning provides no consistent advantage across tasks or architectures, and models exhibit strong sensitivity to prompt formulation. Minor wording changes produce large swings in accuracy and refusal rates. To test whether the closed-form VQA suppresses latent knowledge, we prompt each VLM to produce a free-form description of the image and then use GPT-5.1 to infer a diagnosis from the generated description alone. Improvements are limited and remain bounded by task difficulty. Finally, analysis of vision encoder embeddings shows that disease separability varies across datasets, suggesting that limitations arise both from weak visual representations and downstream reasoning. These results show that apparent gains from medical fine-tuning often reflect task simplicity rather than clinical reasoning, and highlight that medical VLM performance is fragile, prompt-dependent, and not reliably improved by domain-specific fine-tuning.\footnote{Code: \url{https://github.com/michalg04/medical-vlm-fragility/}}

\end{abstract}

\section{Introduction}
\begin{figure}
\centering
\includegraphics[width=0.85\textwidth]{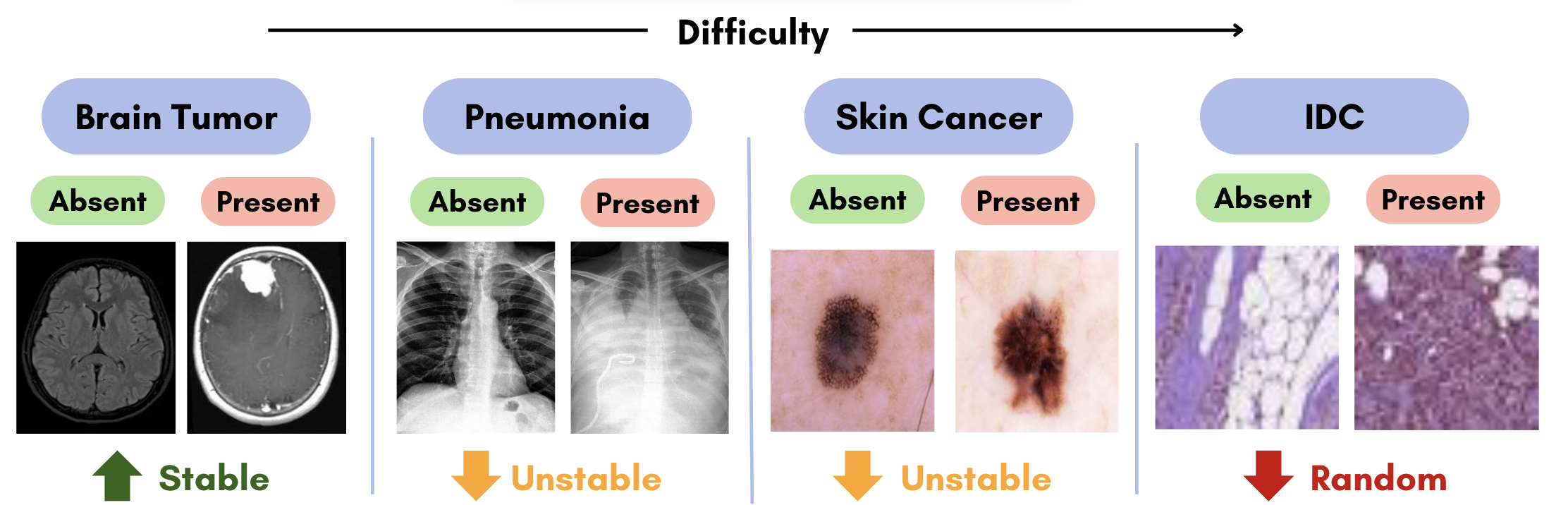}
\caption{Overview of our four tasks. Visually salient tasks (e.g., brain tumor) are reliably solved, while more complex tasks (e.g., pneumonia, skin cancer, Invasive Ductal Carcinoma (IDC)) exhibit instability or near-random performance.}
\label{fig:overview}
\end{figure}
Vision-language models (VLMs) have been adapted for medical applications through domain-specific fine-tuning on radiology reports \cite{goswami2025medivlm}, clinical captions \cite{zhang2023biomedclip}, and biomedical instruction data \cite{cui2024biomedical}. Models such as LLaVA-Med \cite{li2023llava} and MedGemma \cite{sellergren2025medgemma} report improved performance on medical benchmarks, suggesting that domain adaptation may improve clinical reasoning. However, growing evidence suggests that benchmark gains may not reflect robust diagnostic understanding. Models may rely on dataset-specific correlations or superficial visual cues, and exhibit substantial sensitivity to prompt formulation, leading to large performance differentials under semantically equivalent inputs \cite{DeGrave2020,oakdenrayner2019hiddenstratificationcausesclinically,sclar2023quantifying}. 

In this work, we perform a controlled evaluation of base and medically fine-tuned VLMs across four diagnostic tasks: brain tumor identification, pneumonia detection, skin cancer classification, and Invasive Ductal Carcinoma (IDC) classification (shown in Figure \ref{fig:overview}). These tasks span increasing diagnostic complexity, from simple feature identification (brain tumor detection) to integration of subtle visual cues and clinical expertise (IDC classification).
\begin{wrapfigure}{r}{0.38\linewidth}
    \centering
    \includegraphics[width=\linewidth]{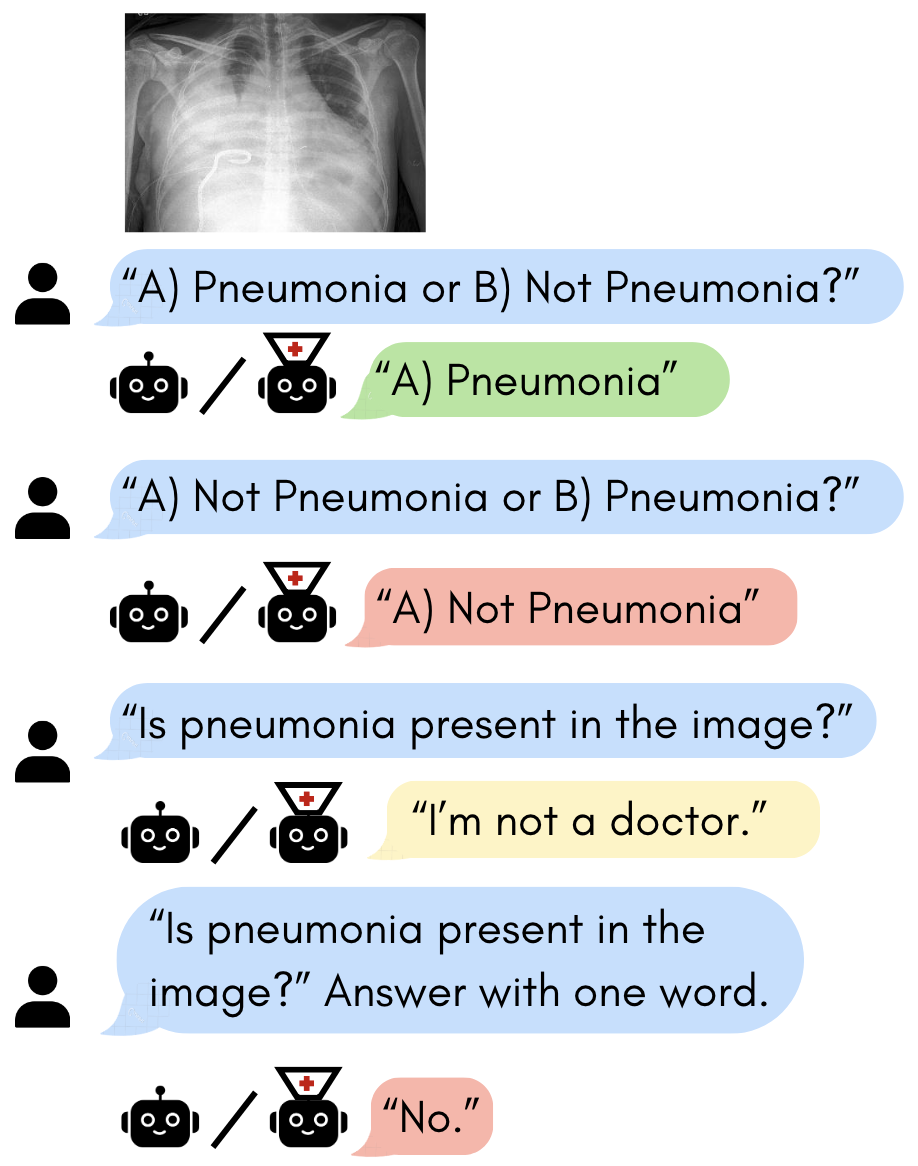}
    \caption{Small prompt changes cause inconsistent decisions in both medical and general VLMs.}
    \label{fig:teaser}
\end{wrapfigure}
Our results show that medical fine-tuning only provides gains on tasks requiring limited clinical knowledge, yet produces random performance on challenging diagnostic tasks, indicating that current medical-VLMs do not possess true clinical reasoning. We first evaluate whether domain-specific fine-tuning translates into improved performance on simple diagnostic visual question answering (VQA) tasks. We create 12 prompt variations spanning multiple-choice (A/B), short-form (yes/no), and free-response, allowing us to test whether similar prompt formulations elicit the same answers as the structured template prompts used during fine-tuning. This yields 19{,}200 total model responses across all image-prompt pairs. On brain tumor detection, most models, including base models without medical training, perform strongly (e.g., Gemma achieves 70\% F1-Score, MedGemma reaches 83\%), indicating that the task can be reasonably completed without medical knowledge. Despite the gains on simple tasks, medical fine-tuning does not reliably improve performance. While MedGemma improves over its base model, LLaVA-Med substantially underperforms LLaVA on the same task (25 vs.\ 66 F1), suggesting that fine-tuning can degrade performance even in tasks with clear visual cues.

As task difficulty increases, performance drops sharply for all models. For pneumonia detection, accuracy varies widely across prompts (45–70\%), while for skin cancer and histopathology, most models operate near chance (50\%) despite LLaVA-Med and MedGemma having domain-specific training. Across these harder tasks, medically adapted models do not show consistent advantages over their base counterparts. 

Beyond overall accuracy, we find that prompt formulation can strongly influence measured performance as shown in Figure \ref{fig:teaser}. Namely, minor, semantically equivalent changes, such as reversing answer order or rephrasing a question, produce swings of up to 20 points in accuracy (e.g., 56\% to 75\% on pneumonia for MedGemma). For LLaVA, we find that without short-format prompt instructions, such as ``answer with one word, refusal rates exceed 70\%. Moreover, large standard deviations (up to $\pm$30 points) further indicate substantial instability over different prompt variations.

These limitations raise the question as to whether closed-form VQA is too restrictive to elicit the model’s underlying knowledge. To understand this, we introduce a description-based pipeline in which models generate free-form image descriptions that are then interpreted by a text-only language model (GPT-5). This recovers partial diagnostic signal (e.g., +10–23 F1 on intermediate tasks), but gains remain bounded by task difficulty and do not resolve failures on the hardest datasets.

Our findings suggest that apparent improvements from medical fine-tuning often reflect task simplicity and evaluation artifacts rather than genuine clinical reasoning. Current medical VLMs remain fragile, highly sensitive to prompt design, and unreliable on clinically realistic tasks, highlighting the need for evaluation protocols that disentangle visual perception, reasoning, and robustness.

\section{Related Work}

\paragraph{Medical vision-language models}
LLaVA-Med \cite{li2023llava} applies curriculum learning on biomedical figure-caption pairs to the LLaVA \cite{liu2023visualinstructiontuning} architecture, reporting gains on medical VQA tasks such as radiology and pathology. MedGemma \cite{sellergren2025medgemma} extends Gemma 3 \cite{kamath2025gemma} through both language and vision-encoder medical adaptation, achieving improvements across multimodal QA, including chest X-ray, histopathology, and dermatology classification. These models are typically evaluated on curated benchmark suites, where improved performance is often interpreted as evidence that medical fine-tuning enhances diagnostic capability. However, this assumption is not consistently supported: \cite{jiang2025hulu} find that LLaVA-Med underperforms general-purpose models on broader evaluations, raising questions about the robustness of these gains.
\paragraph{Benchmark limitations in medical imaging}
A growing body of work shows that strong performance on medical imaging benchmarks does not necessarily reflect clinically meaningful understanding. Models may rely on superficial visual shortcuts rather than pathology, as COVID-19 classifiers have been shown to depend on scanner artifacts \cite{DeGrave2020} and dermoscopy models can be misled by benign skin markings \cite{Winkler2019}. Evaluation can also obscure important failures, as \cite{oakdenrayner2019hiddenstratificationcausesclinically} show that aggregate accuracy may hide poor performance on clinically relevant subgroups. In addition, \cite{asadi2026mirageillusionvisualunderstanding} demonstrates that text-only models can perform well on Chest X-Ray VQA, suggesting that non-visual signals can drive benchmark results. These findings indicate that benchmark performance may be shaped by spurious correlations, evaluation artifacts, and non-visual cues rather than robust diagnostic understanding.
\paragraph{Input sensitivity in LLMs and VLMs}
In parallel, recent work shows that LLMs and VLMs are highly sensitive to prompting and evaluation protocols \cite{biderman2024lessonstrenchesreproducibleevaluation}. Small, meaning-preserving changes in input format can produce large performance differences, up to 76 points in accuracy in few-shot settings \cite{sclar2023quantifying}, while option ordering alone can significantly alter predictions \cite{zheng2024largelanguagemodelsrobust}. In medical VLMs, methods like chain-of-thought prompting exaggerate prompt sensitivity \cite{sadanandan2026chain}. Additionally, \cite{rottger2024xstesttestsuiteidentifying} shows that safety-tuned models may refuse otherwise benign queries depending on prompt phrasing, further highlighting sensitivity to input formulation.

In this work, we study medical and general-purpose VLMs across task difficulties and prompt variation to disentangle how these factors influence measured performance.

\section{Methods}
\label{tasks}

\paragraph{Models.} We evaluate four publicly available vision-language models forming two base/medical pairs: LLaVA and LLaVA-Med, and Gemma and MedGemma, enabling direct comparison between general-purpose and medically fine-tuned models. For the first pair, we use LLaVA v1.6 Mistral 7B \cite{liu2024llavanext} and LLaVA-Med v1.5 Mistral 7B \cite{li2023llava}. For the second pair, we use Gemma-3 4B Instruct \cite{kamath2025gemma}, and MedGemma 4B \cite{sellergren2025medgemma}, a clinically fine-tuned variant of the Gemma-3 base model. All models fall in the 4B--7B parameter range, reflecting a realistic on-premise deployment setting for resource-constrained medical applications.


\paragraph{Tasks.} We evaluate four binary classification tasks spanning common medical imaging datasets: brain tumor identification (MRI), pneumonia detection (chest X-ray), skin cancer classification (dermatoscopy), and histopathology cancer recognition.

\begin{enumerate}
\item \textbf{Brain Tumor Identification (MRI):} We use the BRISC dataset \cite{fateh2026brisc}, which contains radiologist-annotated brain MRIs with and without tumors. Positive samples are scans with tumors and negatives are tumor-free scans. We uniformly sample positives across tumor types and anatomical planes, and sample negatives from the corresponding healthy distribution.

\item \textbf{Pneumonia Detection (Chest X-ray):} We use CheXpert \cite{irvin2019chexpertlargechestradiograph}. Positives are low-uncertainty Pneumonia cases, and negatives are drawn from the ``No Finding'' subset to ensure clear label separation.

\item \textbf{Skin Cancer Classification (Dermatoscopy):} We use dermatoscopic images from the ISIC archive \cite{center2022international}. Positives are melanoma and negatives are benign nevi, sampled to reflect a range of lesion appearances.

\item \textbf{Invasive Ductal Carcinoma (IDC) Classification:} We use the IDC dataset \cite{janowczyk2016deep} containing histopothological cancer images. Positives are patches containing invasive ductal carcinoma, and negatives are non-cancerous tissue.
\end{enumerate}

The tasks are ordered by increasing diagnostic complexity, which we characterize by the length of the implicit diagnostic rubric (i.e., the number of visual criteria required for a decision). Brain tumor detection often reduces to identifying a salient mass, whereas skin cancer and histopathology require evaluating multiple subtle features across regions.

\subsection{Closed-Form VQA Evaluation Protocol}\label{closed-form}

We evaluate whether domain-specific fine-tuning preserves robust visual reasoning, focusing on stability under prompt variation and the ability to produce reliable answers.

A central challenge is non-committal behavior, where models refuse to answer the prompt. To standardize evaluation, we use GPT-5.1 as a VLM judge to map all outputs to one of three categories: \emph{correct}, \emph{incorrect}, or \emph{refused}, where refusals include any response that cannot be unambiguously mapped to a target label (details in Appendix~\ref{judge_prompts}). We report two complementary F1-scores. F1$_{nr}$ is computed over non-refused predictions and reflects performance when the model commits to an answer, while F1$_{all}$ treats refusals as incorrect, capturing effective performance under abstention.


\paragraph{Prompt design.} \label{prompt_design}

To measure prompt sensitivity, we evaluate semantically equivalent queries expressed in multiple forms. For each diagnostic target $d \in \{\text{brain tumor}, \text{pneumonia}, \text{melanoma}, \text{invasive ductal carcinoma}\}$ as defined in Section~\ref{tasks}, we use the following prompt templates:

\textbf{Binary forced choice (A/B):} \textit{Which does this image show? Choose one answer A) $d$ \; B) No $d$}

\textbf{Binary forced choice (B/A):} \textit{Which does this image show? Choose one answer A) No $d$ \; B) $d$}

\textbf{Normality assessment:} \textit{Is this image normal?} \quad \textit{Is this image abnormal?}

\textbf{Direct presence (minimal):} \textit{Does this image contain $d$?}

\textbf{Direct presence (guided):} \textit{Does this image contain $d$? Consider clinical signs $s_1, \dots, s_n$.}

The guided variant includes a short list of clinically relevant signs ($s_1, \dots, s_n$) associated with a positive diagnosis. Full prompt and symptom lists are provided in Appendix \ref{desc:full_list}.

Each prompt is evaluated under two response settings. In the \textbf{constrained} setting, models are instructed to produce a single-word answer (implemented by appending \textit{"Answer in one word"} to the prompt). In the \textbf{open-ended} setting, free-form responses are allowed, exposing tendencies toward hedging or refusal.

For each task, we sample a balanced set of 50 positive and 50 negative images. Each image is evaluated across 4 models, 6 prompt variants, and 2 response settings, yielding 4{,}800 evaluations per task and 19{,}200 total.

\subsection{Open-Ended Medical Description and Information Recovery Protocol} \label{open-ended}


To test whether closed-form prompting obscures clinical reasoning, we intro two-step pipeline where: 1) a medical VLM describes an image in detail, then 2) GPT-5.1 maps the generated description to a diagnosis. If the model’s description contains enough information for a frontier text-only model to recover the correct diagnosis, then the failure lies in how the answer is elicited in VQA rather than in the underlying visual representation.

For each image and diagnostic target $d$, we prompt the model with either a general instruction (\textit{``Describe this image in great detail and pay attention to any relevant pixels. Imagine a physician will use your description.''}) or a diagnosis-oriented variant that conditions on $d$ (\textit{``...Imagine a physician will use your description to diagnose $d$.''}). This allows us to compare unguided descriptions with task-specific, diagnosis-aware ones.

The resulting descriptions are evaluated by GPT-5.1 using prompts of the form \textit{``Based on the following description, is $d$ present? Answer Yes or No.''} The evaluator receives only the description and task instruction, with no access to the original image. We use three prompt variants per disease and query the model zero-shot with default decoding. Outputs are parsed to extract binary predictions, and no refusals are observed. Full evaluator prompts are provided in Appendix \ref{evaluator_prompts}. Performance in our two-step pipeline reflects a VLM's ability to produce clinically relevant image summaries. We treat the evaluator as an approximate upper bound on extractable diagnostic signal from the model’s outputs.


\section{Results}


\subsection{Closed-Form Medical VQA Performance}

\begin{figure*}[t]
\centering
\includegraphics[width=0.6\textwidth]{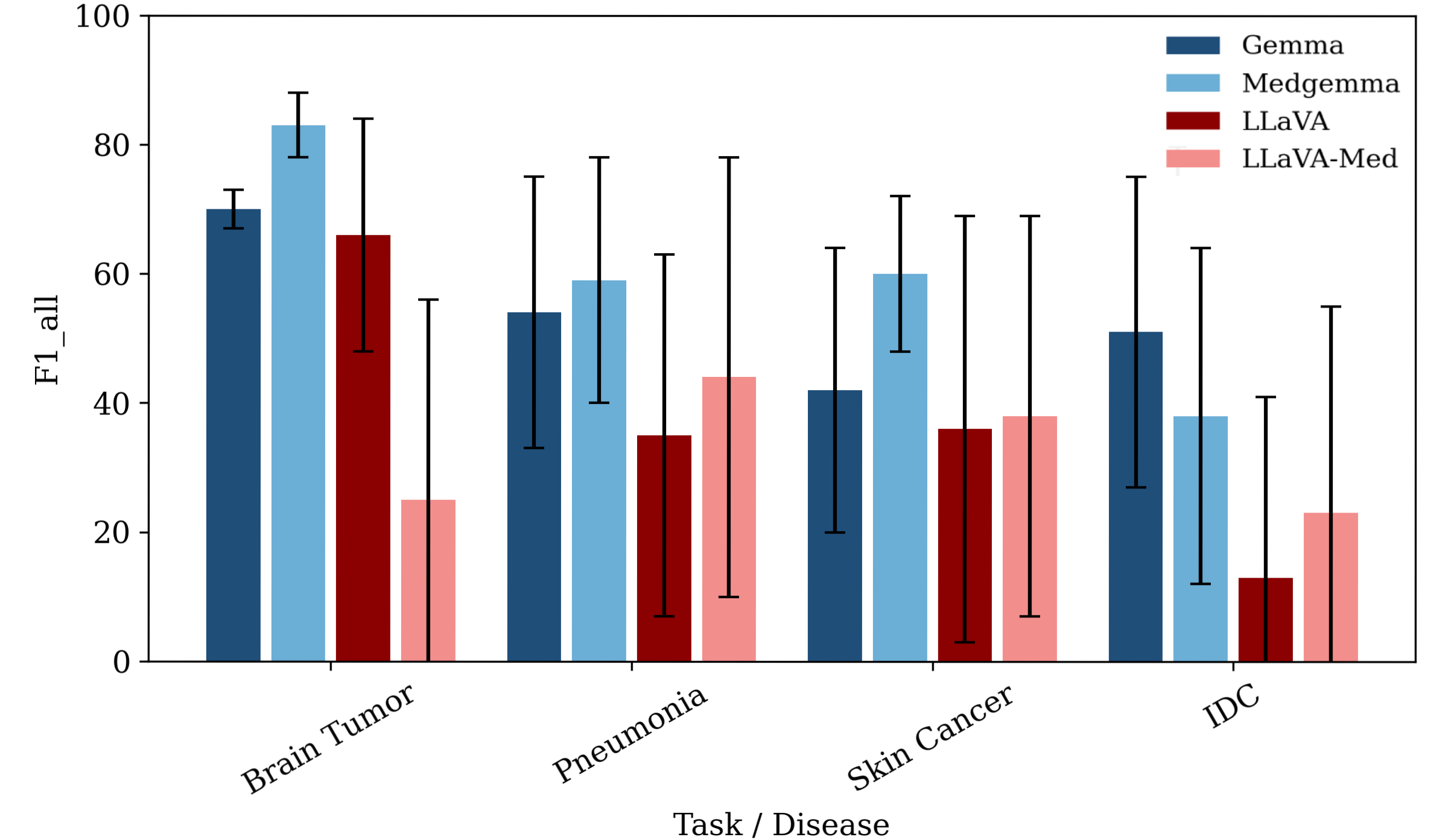}
\caption{
Performance (F1$_{all}$) across disease classification tasks for base and medically fine-tuned models, averaged across all prompt variants. Medical fine-tuning yields improvements on some tasks but does not consistently outperform base models.
}
\label{fig:f1_all}
\end{figure*}

\begin{table}[t]
\centering
\small
\setlength{\tabcolsep}{3pt}
\begin{adjustbox}{scale=0.85}
\begin{tabular}{llccccc}
\toprule
Model & Prompt Variant & Corr $\uparrow$ & Inc $\downarrow$ & Ref $\downarrow$ & F1$_{nr}$ $\uparrow$ & F1$_{all}$ $\uparrow$ \\
\midrule
Gemma 
 & A) Pneumonia \; B) No Pneumonia & 56 & 38 & 6  & 72 & 67 \\
 & A) No Pneumonia \; B) Pneumonia & 52 & 47 & 1  & 68 & 67 \\
 & Is the image normal?           & 56 & 44 & 0  & 69 & 69 \\
 & Is the image abnormal?         & 51 & 39 & 10 & 70 & 63 \\
 & Pneumonia present? (consider ...)  & 41 & 14 & 45 & 81 & 45 \\
 & Pneumonia present?             & 15 & 1  & 84 & 93 & 15 \\
\midrule
MedGemma 
 & A) Pneumonia \; B) No Pneumonia & 56 & 44 & 0 & 21 & 21 \\
 & A) No Pneumonia \; B) Pneumonia & 75 & 25 & 0 & 67 & 67 \\
 & Is the image normal?           & 70 & 30 & 0 & 69 & 69 \\
 & Is the image abnormal?         & 68 & 32 & 0 & 66 & 66 \\
 & \textbf{Pneumonia present? (consider ...)}  & \textbf{76} & \textbf{24} & \textbf{0} & \textbf{69} & \textbf{69} \\
 & Pneumonia present?             & 72 & 27 & 1 & 62 & 61 \\
\midrule
LLaVA 
 & A) Pneumonia \; B) No Pneumonia & 50 & 50 & 0 & 0  & 0  \\
 & A) No Pneumonia \; B) Pneumonia & 60 & 40 & 0 & 66 & 66 \\
 & Is the image normal?           & 55 & 45 & 0 & 53 & 53 \\
 & Is the image abnormal?         & 50 & 50 & 0 & 7  & 7  \\
 & Pneumonia present? (consider ...)  & 55 & 45 & 0 & 26 & 26 \\
 & Pneumonia present?             & 64 & 36 & 0 & 59 & 59 \\
\midrule
LLaVA-Med 
 & A) Pneumonia \; B) No Pneumonia & 50 & 50 & 0 & 0  & 0  \\
 & A) No Pneumonia \; B) Pneumonia & 50 & 50 & 0 & 0  & 0  \\
 & Is the image normal?           & 50 & 50 & 0 & 67 & 67 \\
 & Is the image abnormal?         & 50 & 50 & 0 & 67 & 67 \\
 & Pneumonia present? (consider ...)  & 50 & 50 & 0 & 67 & 67 \\
 & Pneumonia present?             & 50 & 50 & 0 & 67 & 67 \\
\bottomrule
\end{tabular}
\end{adjustbox}
\caption{Closed-form VQA prompt sensitivity results on pneumonia detection. Corr = correct, Inc = incorrect, Ref = refused.}
\label{tab:pneumonia-oneword}
\end{table}

\paragraph{Clinical reasoning quality degrades with task difficulty.}
Figure \ref{fig:f1_all} reports F1$_{all}$ results averaged across all prompt variants described in Section \ref{prompt_design}. This metric, defined in Section \ref{closed-form}, provides a more stable estimate of model behavior than any single formulation. Prompts include an appended instruction to \emph{answer with one word}, forcing discrete outputs (e.g., ``pneumonia,'' ``normal'') and reducing ambiguity in evaluation. Even under this constraint, performance varies substantially across tasks and prompt variants, indicating inherent instability. Full results for all metrics and prompts are provided in Appendix \ref{app:full_results}.

Brain tumor identification is consistently the easiest task. All models perform well above chance, with MedGemma reaching 83 F1$_{all}$, and base models (Gemma, LLaVA) exceeding 60\%. This suggests that success is driven by the presence of clear visual features, as brain tumors often appear as high-contrast, spatially distinct regions in MRI \cite{martucci2023magnetic}, rather than requiring specialized reasoning. Medical fine-tuning does produce performance gains, with MedGemma outperforming its base counterpart. However, this benefit is not consistent, as LLaVA-Med underperforms compared to LLaVA, suggesting that fine-tuning may degrade general visual processing in some cases.

Performance drops sharply on pneumonia, IDC, and skin cancer. For pneumonia, F1 ranges from 35\% (LLaVA) to 59\% (MedGemma), with large variability (up to, $\pm 34$), indicating sensitivity to prompt formulation. On IDC and skin cancer, most models operate near or below random chance ($\sim$50\%), with some substantially worse (e.g., LLaVA: 13\% on IDC). These results are notable given that these domains overlap with the medical training data, suggesting that fine-tuning does not reliably translate into robust diagnostic performance. In particular, MedGemma and LLaVA-Med were both trained on MRI, CT, and histopathology images, and MedGemma was additionally trained on dermatology images. 


\paragraph{Medical fine-tuning provides no consistent benefits.}
Comparisons between base and medically fine-tuned models show no consistent advantage of medical fine-tuning. While MedGemma performs strongly on brain tumor detection (83\% F1), this does not generalize. In particular, MedGemma's performance on IDC remains near chance (38\% F1), while Gemma achieves a higher F1 despite no medical training. LLaVA and LLaVA-Med also diverge on easier tasks (brain tumor) but converge on harder ones (skin cancer), suggesting that fine-tuning does not systematically improve robustness.
\paragraph{Prediction collapse and calibration failures.}
A recurring failure mode is prediction collapse, where models default to a single class regardless of input. This appears as near-chance accuracy with low or highly variable F1 (e.g., LLaVA on IDC: 13 $\pm$ 28), reflecting poor calibration and weak sensitivity to image content. Aggregate accuracy obscures these behaviors, highlighting the need to consider class balance, refusal, and variability across prompts.

\paragraph{Prompt sensitivity and evaluation fragility.} 
We next examine the effects of prompt variation on a single representative task. Our prompt variations are intentionally simpler than those used in medical fine-tuning of LLaVA-Med and MedGemma, allowing us to test robustness under semantically equivalent prompts. We focus on pneumonia, since unlike brain tumor identification, it does not admit obvious visual cues, but it also does not collapse to near-random performance as in IDC. Results for other tasks are in Appendix \ref{app:prompts_var}.

Table~\ref{tab:pneumonia-oneword} shows performance varies substantially across minor prompt changes. Surface-level modification, including answer-order reversal, semantic framing (``normal'' vs.\ ``abnormal''), and whether the model is asked to directly commit to ``pneumonia present'', produce large swings in measured performance. For example, MedGemma increases from 56\% to 75\% correct simply by reversing the answer order (\emph{A) Pneumonia, B) No Pneumonia} vs.\ \emph{A) No Pneumonia, B) Pneumonia}), and reaches 76\% on \emph{Pneumonia present? (consider signs of ...)}. Gemma shows a different instability: it achieves 56\% on both answer-order and ``normal'' prompts, but drops to 41\% on \emph{Pneumonia present? (consider ...)} and 15\% on the direct \emph{Pneumonia present?} formulation. LLaVA varies from 50\% to 64\% across prompts, while LLaVA-Med remains flat at 50\% throughout. LLaVA-Med performance is concerning but aligns with recent work, which also shows suboptimal results \cite{jiang2025hulu}. 

Refusal behavior further amplifies these differences. For Gemma, refusals increase from 1\% on the answer-order prompt to 45\% and 84\% on the two \emph{Pneumonia present?} variants, indicating that some prompt formulations elicit avoidance rather than incorrect predictions. Overall, small wording changes can dominate measured performance, and medically adapted models are not necessarily less prompt-sensitive than general-purpose ones. The extreme prompt sensitivity is particularly concerning, as VQA-style prompting reflects real-world deployment, yet we show that VQA may suppress or fail to elicit clinical knowledge.

\paragraph{One-word vs.\ unconstrained prompting.}
\begin{table}[t]
\centering
\small
\setlength{\tabcolsep}{4pt}
\begin{tabular}{llccccc}
\toprule
Model & Setting & Corr $\uparrow$ & Inc $\downarrow$ & Ref $\downarrow$ & F1$_{nr}$ $\uparrow$ & F1$_{all}$ $\uparrow$ \\
\midrule
Gemma    
 & With one word    & 45 & 34 & 22 & 71 & 54 \\
 & Without one word & 45 & 35 & 20 & 68 & 62 \\
\midrule
MedGemma 
 & With one word    & 68 & 32 & 1  & 60 & 59 \\
 & Without one word & 36 & 27 & 38 & 56 & 42 \\
\midrule
LLaVA    
 & With one word    & 58 & 42 & 0  & 38 & 38 \\
 & Without one word & 17 & 11 & 72 & 19 & 12 \\
\midrule
LLaVA-Med 
 & With one word    & 52 & 48 & 0  & 38 & 38 \\
 & Without one word & 47 & 42 & 11 & 32 & 31 \\
\bottomrule
\end{tabular}
\caption{Average performance across all diseases (brain tumor, pneumonia, IDC, and skin cancer), comparing one-word and unconstrained VQA.}
\label{tab:all-disease-avg-compare}
\end{table}
Up to this point, we have focused on the \emph{answer with one word} setting, which enforces a constrained format and yields more consistent outputs. We now compare against the same prompts without this constraint. The most striking change is a sharp increase in refusal rates, as removing the one-word requirement leads many models to produce safety-oriented responses instead of committing to an answer. This suggests that the constraint suppresses such behavior. Gemma is an exception, showing relatively consistent behavior across both settings.

Table~\ref{tab:all-disease-avg-compare} summarizes average performance across all diseases and prompt variants. Constraining the response format substantially affects both performance and refusal behavior. MedGemma provides the clearest example. Average correctness increases from 36\% to 68\%, refusals drop from 38\% to 1\%, and $F1_{all}$ improves from 42\% to 59\%. LLaVA shows an even stronger dependence. Without the constraint, it achieves 17\% correctness with 72\% refusals and $F1_{all}=12\%$. With the constraint, correctness rises to 58\%, refusals drop to 0\%, and $F1_{all}$ increases to 38\%. These effects are not apparent from any single task or prompt, but emerge in the aggregate. Overall, unconstrained prompting allows models to hedge or avoid committing to an answer, while the one-word setting forces a discrete diagnostic decision and makes both capability and refusal behavior more directly observable.


\subsection{Beyond VQA: Description-Based Evaluation Partially Recovers Diagnostic Signal}\label{beyond_vqa}

\begin{table*}[t]
\centering
\small
\setlength{\tabcolsep}{4pt}
\begin{tabular}{lcccccccc}
\toprule
& \multicolumn{2}{c}{Brain Tumor} 
& \multicolumn{2}{c}{Pneumonia} 
& \multicolumn{2}{c}{Skin Cancer} 
& \multicolumn{2}{c}{IDC} \\
\cmidrule(lr){2-3} \cmidrule(lr){4-5} \cmidrule(lr){6-7} \cmidrule(lr){8-9}
Model 
& Corr $\uparrow$ & F1 $\uparrow$ 
& Corr $\uparrow$ & F1 $\uparrow$ 
& Corr $\uparrow$ & F1 $\uparrow$ 
& Corr $\uparrow$ & F1 $\uparrow$ \\
\midrule
Gemma 
& 76 {\scriptsize $\pm$ 8}  & 80 {\scriptsize $\pm$ 5}
& 54 {\scriptsize $\pm$ 1}  & \textbf{65 {\scriptsize $\pm$ 5}}
& 50 {\scriptsize $\pm$ 2}  & 65 {\scriptsize $\pm$ 1}
& 52 {\scriptsize $\pm$ 4}  & 45 {\scriptsize $\pm$ 29} \\

MedGemma 
& \textbf{82 {\scriptsize $\pm$ 5}}  & \textbf{84 {\scriptsize $\pm$ 3}}
& \textbf{70 {\scriptsize $\pm$ 8}}  & 64 {\scriptsize $\pm$ 22}
& \textbf{55 {\scriptsize $\pm$ 6}}  & \textbf{67 {\scriptsize $\pm$ 2}}
& 51 {\scriptsize $\pm$ 2}  & 43 {\scriptsize $\pm$ 29} \\

LLaVA 
& 70 {\scriptsize $\pm$ 3}  & 66 {\scriptsize $\pm$ 9}
& 52 {\scriptsize $\pm$ 3}  & 35 {\scriptsize $\pm$ 15}
& 50 {\scriptsize $\pm$ 4}  & 50 {\scriptsize $\pm$ 10}
& 51 {\scriptsize $\pm$ 3}  & 40 {\scriptsize $\pm$ 27} \\

LLaVA-Med 
& 73 {\scriptsize $\pm$ 18} & 78 {\scriptsize $\pm$ 11}
& 51 {\scriptsize $\pm$ 3}  & 45 {\scriptsize $\pm$ 30}
& 42 {\scriptsize $\pm$ 5}  & 42 {\scriptsize $\pm$ 10}
& \textbf{53 {\scriptsize $\pm$ 3}}  & \textbf{50 {\scriptsize $\pm$ 22}} \\
\bottomrule
\end{tabular}
\caption{Results from description-based evaluation, where GPT-5.1 diagnoses diseases using VLM-generated descriptions.}
\label{tab:gpt-pipeline-avg}
\end{table*}

To test whether closed-form VQA underestimates model knowledge, we shift from direct question answering to a two-stage setup. VLMs first produce open-ended descriptions of each image, which are then passed to a text-only model (GPT-5.1) to generate the final prediction. This tests whether a diagnostic signal is present in the visual representations but not expressed in VQA outputs.

Results are shown in Table~\ref{tab:gpt-pipeline-avg}. Across most tasks, this pipeline improves stability and overall performance, with no refusals and higher F1 scores than direct VQA. Gains are largest on intermediate tasks. For pneumonia, MedGemma reaches 64\% F1 and Gemma 65\% F1, about 10\% higher than VQA. For skin cancer, Gemma reaches 65\% F1, a 23\% improvement, suggesting that descriptions capture partially useful diagnostic structure.

However, improvements are not uniform. On brain tumor identification, gains are small (e.g., MedGemma 83 to 84 F1), indicating that VQA already captures most of the signal for visually salient tasks. On IDC, performance remains near chance across models, suggesting little recoverable diagnostic information even in descriptions. Variability remains high in several settings, showing that prompt sensitivity persists. While this approach reduces refusal and improves average performance, it does not eliminate instability. Overall, open-ended descriptions partially recover the diagnostic signal suppressed in VQA, especially for moderately difficult tasks. However, performance remains bounded by task difficulty. When visual cues are highly salient or require complex domain knowledge, the prompting strategy has limited impact. Medical fine-tuning still provides the strongest performance overall, but its advantage is inconsistent across tasks.

\subsection{Visual Separability in Vision Encoder Representations}

\begin{figure*}[t]
\centering
\includegraphics[width=0.7\textwidth]{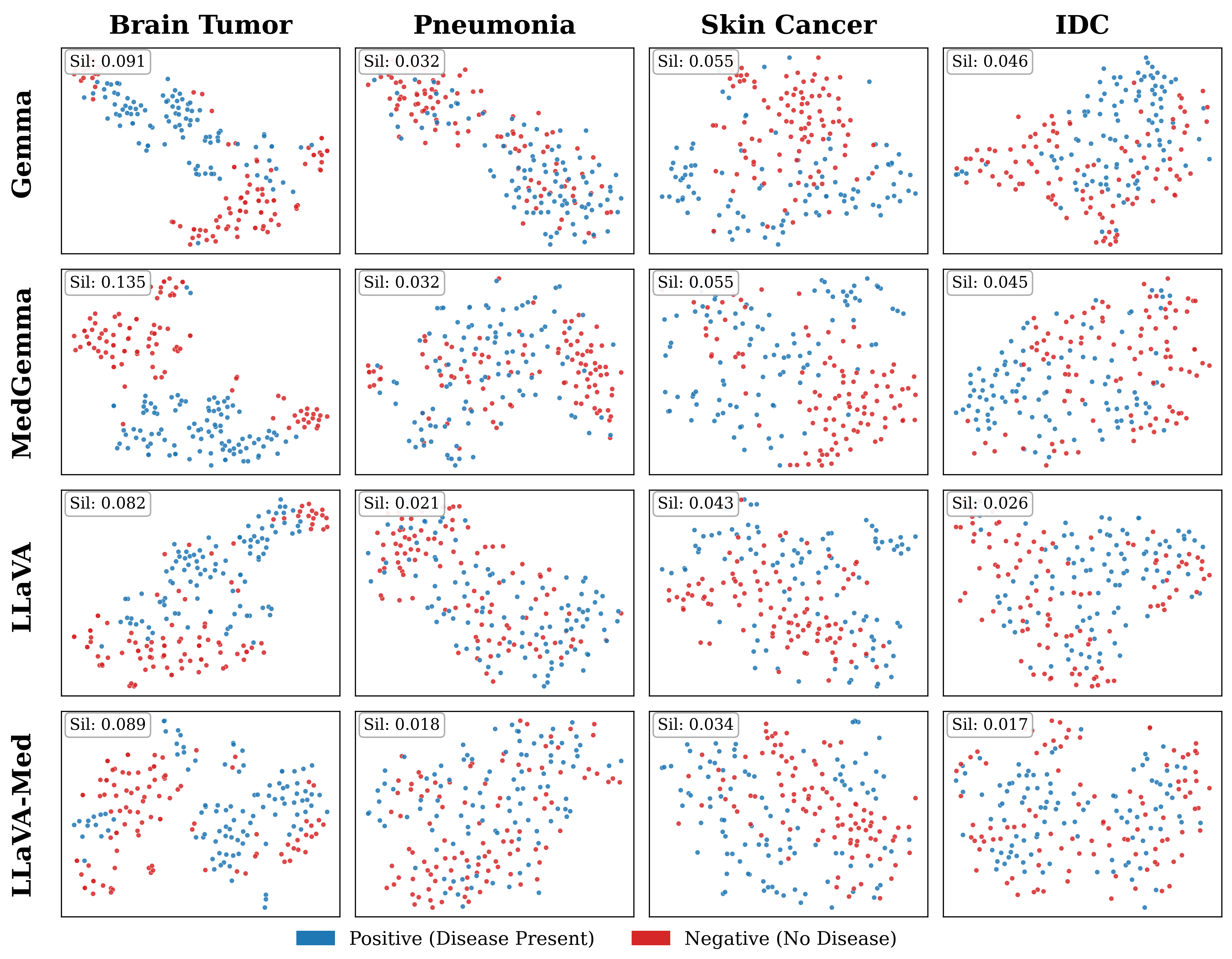}
\caption{t-SNE projections of vision encoder embeddings of the four datasets. Each point represents an image embedding, colored by ground-truth label. Silhouette scores are reported in each subplot to quantify class separability.}
\label{fig:tsne-embeddings}
\end{figure*}

To examine whether failures originate from weak visual representations or downstream reasoning (following \cite{rudman2025forgotten}), we visualize vision encoder embeddings using t-SNE (Figure~\ref{fig:tsne-embeddings}). Each point corresponds to an image embedding colored by its ground-truth label. We quantify class separability using the silhouette score, where values near 1 indicate well-separated clusters and values near 0 indicate overlapping classes.

Brain tumor MRI images show the strongest separation across models (with average silhouette scores of 0.10), indicating distinct clustering of tumor and non-tumor scans. This is consistent with the task’s clear visual features, such as bright, localized masses \cite{martucci2023magnetic}. In contrast, chest X-rays exhibit little separability, with silhouette scores of 0.026, suggesting weak encoder-level signal. Skin cancer images show intermediate structure, with relatively strong clustering despite requiring domain expertise, likely due to visually salient patterns such as asymmetry and color variation. Model differences are also evident. LLaVA and LLaVA-Med show weak clustering on brain tumor images, which may contribute to downstream refusal behavior. More broadly, LLaVA-Med does not consistently improve separability over LLaVA. 

These results suggest that limitations arise at multiple stages of the multi-modal integration. Some tasks are visually separable at the representation level, while others exhibit weak encoder-level signal, helping explain persistent failures even under improved prompting.

\section{Discussion}
Our results expose a critical mismatch between previously reported benchmark performance and the practical reliability of medical vision-language models. Although models can achieve competitive scores on simple medical tasks, these gains are highly sensitive to prompt formulation and do not reflect stable clinical understanding. Minor, semantically equivalent changes, such as reversing multiple-choice order, can induce performance changes of up to 20 points, revealing that reported improvements are often artifacts of evaluation design rather than robustness. Notably, medical fine-tuning does not resolve this fragility and can even amplify it, introducing inconsistent calibration and prompt-dependent refusal behavior. In practice, these models risk deployment by requiring iterative prompt refinement to obtain usable outputs, effectively shifting complexity from model design to the end user. We argue that the community must move beyond single-metric benchmark reporting toward robustness-centered evaluation, including sensitivity to prompt variation, consistency across equivalent inputs, and reliability under realistic interaction settings.

\section{Conclusion}
In this work, we evaluate whether medical fine-tuning improves the robustness and clinical reasoning ability of vision-language models. Across paired base and medically adapted models, we find that perceived gains are limited to simpler diagnostic tasks and prompt formulation rather than medical fine-tuning. While all models perform well on visually straightforward tasks, performance degrades sharply on tasks requiring clinical knowledge, often approaching chance. We further show that evaluation design plays a critical role, with minor prompt variations producing large changes in accuracy and refusal behavior. Open-ended description-based evaluation recovers only a limited additional clinical signal, as opposed to multiple-choice VQA. Analysis of vision encoder representations suggests that these limitations arise from both weak visual separability and downstream reasoning failures. Overall, our results indicate that apparent gains from domain-specific fine-tuning may overstate true progress toward reliable medical reasoning, highlighting the need for more robust evaluation protocols and training methods that yield stable and generalizable multimodal understanding in high-stakes domains.




\bibliography{colm2026_conference}
\bibliographystyle{colm2026_conference}

\appendix

\newpage
\section{Full List of Closed-Form VQA Prompts} \label{desc:full_list}
\begin{table}[h]
\centering
\small
\setlength{\tabcolsep}{4pt}
\begin{tabular}{lp{9cm}}
\toprule
Diagnostic Target & Prompt \\
\midrule
Pneumonia
 & Is this image abnormal? \\[4pt]
 & Is this image normal? \\[4pt]
 & Does this image contain pneumonia? \\[4pt]
 & Does this image contain pneumonia? Consider signs of consolidation, air bronchograms, and silhouettes. \\[4pt]
 & Which does this image show? Choose one answer. A) Pneumonia \; B) No Pneumonia \\[4pt]
 & Which does this image show? Choose one answer. A) No Pneumonia \; B) Pneumonia \\
\midrule
Brain Tumor
 & Is this image abnormal? \\[4pt]
 & Is this image normal? \\[4pt]
 & Does this image contain evidence of tumors? \\[4pt]
 & Does this image contain evidence of tumors? Consider signs of irregular masses and other atypical morphology. \\[4pt]
 & Which does this image show? Choose one answer. A) Tumor \; B) No Tumor \\[4pt]
 & Which does this image show? Choose one answer. A) No Tumor \; B) Tumor \\
\midrule
Skin Cancer
 & Is this image abnormal? \\[4pt]
 & Is this image normal? \\[4pt]
 & Does this image contain evidence of malignant skin cancer? \\[4pt]
 & Does this image contain evidence of malignant skin cancer? Consider signs of irregular borders, asymmetry, uneven pigmentation, and other atypical morphology. \\[4pt]
 & Which does this image show? Choose one answer. A) Malignant \; B) Benign \\[4pt]
 & Which does this image show? Choose one answer. A) Benign \; B) Malignant \\
\midrule
IDC
 & Is this image abnormal? \\[4pt]
 & Is this image normal? \\[4pt]
 & Does this image contain evidence of invasive ductal carcinoma? \\[4pt]
 & Does this image contain evidence of invasive ductal carcinoma? Consider signs of irregular masses and other atypical cell morphology. \\[4pt]
 & Which does this image show? Choose one answer. A) Carcinoma \; B) No Carcinoma \\[4pt]
 & Which does this image show? Choose one answer. A) No Carcinoma \; B) Carcinoma \\
\bottomrule
\end{tabular}
\caption{Prompts used for each disease classification task.}
\label{tab:prompts}
\end{table}

\newpage

\section{Judge Prompts} \label{judge_prompts}

The following system prompts are used by GPT-5.1 to classify VLM responses as correct, incorrect, or refused (AMBIGUOUS), as regular expressions (regex) did not capture enough nuance in responses. Each is paired with a user template containing the original VLM prompt and model response.

\paragraph{Brain Tumor.}
\begin{quote}
\small
You are a strict binary classifier. You will be given a response from a vision-language model that was shown a brain MRI image and asked a question about it.

Your job: determine whether the model's response indicates the image shows a TUMOR (i.e., contains a brain tumor / is abnormal / shows a mass or lesion) or NO\_TUMOR (i.e., no tumor / normal / healthy brain).

Rules:

-- If the model says the image IS abnormal, DOES contain a tumor, shows a mass/lesion, answers ``A) Brain tumor'', or otherwise indicates a tumor is present then respond exactly: TUMOR

-- If the model says the image is normal, healthy, no tumor, no mass, no abnormality, answers ``B) No brain tumor'', or otherwise indicates no tumor then respond exactly: NO\_TUMOR

-- If the response is truly ambiguous, hedged equally, or refuses to answer then respond exactly: AMBIGUOUS

You must respond with exactly one word: TUMOR, NO\_TUMOR, or AMBIGUOUS. Nothing else.
\end{quote}

\paragraph{Pneumonia.}
\begin{quote}
\small
You are a strict binary classifier. You will be given a response from a vision-language model that was shown a chest X-ray and asked a question about it.

Your job: determine whether the model's response indicates the image is POSITIVE (i.e., contains pneumonia / is abnormal / shows evidence of pneumonia) or NEGATIVE (i.e., no pneumonia / normal / benign).

Rules:

-- If the model says the X-ray IS abnormal, HAS Pneumonia, DOES contain evidence of pneumonia, answers ``A) Pneumonia'' or ``B) Pneumonia'' (depending on answer order), or otherwise indicates a positive diagnosis then respond exactly: POSITIVE

-- If the model says the X-ray is normal, negative, does NOT contain evidence of pneumonia, answers ``A) No Pneumonia'' or ``B) No Pneumonia'' (depending on answer order), or otherwise indicates NEGATIVE then respond exactly: NEGATIVE

-- If the response is truly ambiguous, hedged equally, or refuses to answer then respond exactly: AMBIGUOUS

You must respond with exactly one word: POSITIVE, NEGATIVE, or AMBIGUOUS. Nothing else.
\end{quote}

\paragraph{Skin Cancer.}
\begin{quote}
\small
You are a strict binary classifier. You will be given a response from a vision-language model that was shown a skin lesion image and asked a question about it.

Your job: determine whether the model's response indicates the image is MALIGNANT (i.e., contains skin cancer / is abnormal / shows malignancy) or BENIGN (i.e., no skin cancer / normal / benign).

Rules:

-- If the model says the lesion IS abnormal, IS malignant, DOES contain evidence of malignant skin cancer, answers ``A) Malignant'' or ``B) Malignant'' (depending on answer order), or otherwise indicates malignancy then respond exactly: MALIGNANT

-- If the model says the lesion is normal, benign, does NOT contain evidence of malignant skin cancer, answers ``A) Benign'' or ``B) Benign'' (depending on answer order), or otherwise indicates benign then respond exactly: BENIGN

-- If the response is truly ambiguous, hedged equally, or refuses to answer then respond exactly: AMBIGUOUS

You must respond with exactly one word: MALIGNANT, BENIGN, or AMBIGUOUS. Nothing else.
\end{quote}

\paragraph{IDC.}
\begin{quote}
\small
You are a strict binary classifier. You will be given a response from a vision-language model that was shown a breast histopathology image and asked a question about it.

Your job: determine whether the model's response indicates the image is MALIGNANT (i.e., contains invasive ductal carcinoma / is abnormal / shows carcinoma) or BENIGN (i.e., no carcinoma / normal / benign).

Rules:

-- If the model says the image IS abnormal, IS carcinoma, DOES contain IDC, answers ``A) Carcinoma'' or ``B) Carcinoma'' (depending on answer order), or otherwise indicates malignancy then respond exactly: MALIGNANT

-- If the model says the image is normal, benign, no carcinoma, does NOT contain IDC, answers ``A) No carcinoma'' or ``B) No carcinoma'' (depending on answer order), or otherwise indicates benign then respond exactly: BENIGN

-- If the response is truly ambiguous, hedged equally, or refuses to answer then respond exactly: AMBIGUOUS

You must respond with exactly one word: MALIGNANT, BENIGN, or AMBIGUOUS. Nothing else.
\end{quote}

\paragraph{User template (shared across all tasks).}
\begin{quote}
\small
Original prompt given to the model:

\{prompt\}

Model's response:

\{response\}

Classification (one word):
\end{quote}

\section{Description-Based Evaluation Full Prompts}\label{evaluator_prompts}
Table \ref{tab:gpt-prompts} shows the full prompts shown to GPT-5.1 to achieve a classification from the VLM outputs. This corresponds to the results in Section \ref{beyond_vqa}.
\begin{table*}[h]
\centering
\small
\setlength{\tabcolsep}{4pt}
\begin{tabular}{llp{9cm}}
\toprule
Diagnostic Target & Variant & Prompt \\
\midrule
\multirow{3}{*}{Brain Tumor}
 & Role-based & You are a careful radiologist. Given the following brain MRI image description, answer ONLY whether a brain tumor is present. Return a single word: Yes or No. \\[4pt]
 & Minimal & Task: Decide if a brain tumor is present. Output must be Yes or No. \\[4pt]
 & Disease-framed & Task: Decide if malignancy is present. Output must be Yes or No. \\
\midrule
\multirow{3}{*}{Pneumonia}
 & Role-based & You are a careful clinician. Given the following lung X-ray description, answer ONLY whether pneumonia is present. Return a single word: Yes or No. \\[4pt]
 & Minimal & Task: Decide if pneumonia is present. Output must be Yes or No. \\[4pt]
 & Disease-framed & Task: Decide if disease is present. Output must be Yes or No. \\
\midrule
\multirow{3}{*}{Skin Cancer}
 & Role-based & You are a careful dermatologist. Given the following skin lesion image description, answer ONLY whether skin cancer is present. Return a single word: Yes or No. \\[4pt]
 & Minimal & Task: Decide if skin cancer is present. Output must be Yes or No. \\[4pt]
 & Disease-framed & Task: Decide if malignancy is present. Output must be Yes or No. \\
\midrule
\multirow{3}{*}{IDC}
 & Role-based & You are a careful histopathologist. Given the following image description, answer ONLY whether inter-ductal carcinoma is present. Return a single word: Yes or No. \\[4pt]
 & Minimal & Task: Decide if inter-ductal carcinoma is present. Output must be Yes or No. \\[4pt]
 & Disease-framed & Task: Decide if malignancy is present. Output must be Yes or No. \\
\bottomrule
\end{tabular}
\caption{GPT-5.1 evaluator prompt variants used in the description-based pipeline. Each prompt is prepended to the VLM-generated description. All variants require a binary Yes/No response.}
\label{tab:gpt-prompts}
\end{table*}

\section{Shared experimental setup}
Across all experiments (both closed-form VQA and description-based evaluation), the same 100 samples described in Section \ref{closed-form} are held fixed across all models, prompt variants, and response settings. All models are run with greedy decoding (temperature 0) and a maximum new token length of 2,048. The description-based pipeline yields an additional 9,600 evaluations, for a combined total of 28,800 across all experiments.

\section{Closed-Form Medical VQA: Additional Results}
\subsection{Results Averaged Across All Datasets}\label{app:full_results}

\begin{table*}[t]
\centering
\small
\setlength{\tabcolsep}{4pt}
\begin{tabular}{llccccc}
\toprule
Disease & Model & Corr & Inc & Ref & F1$_{nr}$ & F1$_{all}$ \\
\midrule
\multirow{4}{*}{Brain Tumor}
& Gemma    & 55.5 $\pm$ 4.8  & 39.0 $\pm$ 12.0 & 5.5 $\pm$ 7.8  & 72.1 $\pm$ 6.5  & 67.7 $\pm$ 1.5 \\
& MedGemma & 76.3 $\pm$ 8.6  & 15.8 $\pm$ 5.3  & 7.8 $\pm$ 7.4  & 86.4 $\pm$ 3.9  & 79.6 $\pm$ 7.0 \\
& LLaVA    & 26.3 $\pm$ 27.6 & 7.2 $\pm$ 10.0  & 66.5 $\pm$ 37.5 & 88.5 $\pm$ 13.9 & 28.0 $\pm$ 30.3 \\
& LLaVA-Med & 59.7 $\pm$ 10.0 & 39.0 $\pm$ 8.9  & 1.3 $\pm$ 3.3  & 33.2 $\pm$ 33.0 & 33.2 $\pm$ 33.0 \\
\midrule
\multirow{4}{*}{Pneumonia}
& Gemma    & 51.8 $\pm$ 1.7  & 48.0 $\pm$ 1.8  & 0.2 $\pm$ 0.4  & 67.5 $\pm$ 0.8  & 67.4 $\pm$ 0.8 \\
& MedGemma & 43.2 $\pm$ 23.5 & 30.0 $\pm$ 19.5 & 26.8 $\pm$ 39.2 & 45.7 $\pm$ 25.5 & 39.0 $\pm$ 21.1 \\
& LLaVA    & 16.7 $\pm$ 25.8 & 15.0 $\pm$ 23.2 & 68.3 $\pm$ 49.1 & 0.0 $\pm$ 0.0   & 0.0 $\pm$ 0.0 \\
& LLaVA-Med & 50.7 $\pm$ 1.0  & 49.3 $\pm$ 1.0  & 0.0 $\pm$ 0.0  & 35.9 $\pm$ 34.5 & 35.9 $\pm$ 34.5 \\
\midrule
\multirow{4}{*}{IDC}
& Gemma    & 35.2 $\pm$ 17.2 & 29.3 $\pm$ 13.0 & 35.5 $\pm$ 29.5 & 64.6 $\pm$ 4.5  & 42.5 $\pm$ 21.3 \\
& MedGemma & 23.8 $\pm$ 22.2 & 20.8 $\pm$ 22.6 & 55.3 $\pm$ 44.8 & 60.2 $\pm$ 37.3 & 16.2 $\pm$ 18.9 \\
& LLaVA    & 11.3 $\pm$ 9.2  & 12.8 $\pm$ 11.5 & 75.8 $\pm$ 20.3 & 30.8 $\pm$ 32.1 & 11.0 $\pm$ 13.8 \\
& LLaVA-Med & 30.2 $\pm$ 24.8 & 28.0 $\pm$ 22.5 & 41.8 $\pm$ 46.6 & 12.7 $\pm$ 24.6 & 11.8 $\pm$ 23.5 \\
\midrule
\multirow{4}{*}{Skin Cancer}
& Gemma    & 44.2 $\pm$ 5.0  & 45.0 $\pm$ 7.9  & 10.8 $\pm$ 12.6 & 64.4 $\pm$ 4.1  & 57.2 $\pm$ 6.4 \\
& MedGemma & 31.7 $\pm$ 17.5 & 15.8 $\pm$ 11.1 & 52.5 $\pm$ 28.2 & 77.7 $\pm$ 14.6 & 34.2 $\pm$ 18.7 \\
& LLaVA    & 11.7 $\pm$ 11.7 & 10.8 $\pm$ 11.7 & 77.5 $\pm$ 23.0 & 39.8 $\pm$ 31.7 & 10.4 $\pm$ 16.9 \\
& LLaVA-Med & 47.5 $\pm$ 4.0  & 51.7 $\pm$ 3.0  & 0.8 $\pm$ 1.6  & 44.5 $\pm$ 25.8 & 44.1 $\pm$ 25.6 \\
\bottomrule
\end{tabular}
\caption{Average performance across prompts for the \textbf{without answer in one word} setting. Values are reported as mean $\pm$ standard deviation across prompts. Corr = correct, Inc = incorrect, Ref = refused.}
\label{tab:all-diseases-open}
\end{table*}

\begin{table*}[t]
\centering
\small
\setlength{\tabcolsep}{4pt}
\begin{tabular}{llccccc}
\toprule
Disease & Model & Corr $\uparrow$ & Inc $\downarrow$ & Ref $\downarrow$ & F1$_{nr}$ $\uparrow$ & F1$_{all}$ $\uparrow$ \\
\midrule
\multirow{4}{*}{Brain Tumor}
& Gemma    & 60 {\scriptsize $\pm$ 8}  & 35 {\scriptsize $\pm$ 14} & 6 {\scriptsize $\pm$ 7}  & 75 {\scriptsize $\pm$ 8}  & 70 {\scriptsize $\pm$ 3} \\
& \textbf{MedGemma} & \textbf{79 {\scriptsize $\pm$ 7}}  & \textbf{21 {\scriptsize $\pm$ 7}}  & \textbf{0 {\scriptsize $\pm$ 0}}  & \textbf{83 {\scriptsize $\pm$ 5}}  & \textbf{83 {\scriptsize $\pm$ 5}} \\
& LLaVA    & 69 {\scriptsize $\pm$ 13} & 31 {\scriptsize $\pm$ 13} & 0 {\scriptsize $\pm$ 0}  & 66 {\scriptsize $\pm$ 18} & 66 {\scriptsize $\pm$ 18} \\
& LLaVA-Med & 58 {\scriptsize $\pm$ 10} & 42 {\scriptsize $\pm$ 10} & 0 {\scriptsize $\pm$ 0}  & 25 {\scriptsize $\pm$ 31} & 25 {\scriptsize $\pm$ 31} \\
\midrule
\multirow{4}{*}{Pneumonia}
& Gemma    & 45 {\scriptsize $\pm$ 16} & 31 {\scriptsize $\pm$ 19} & 24 {\scriptsize $\pm$ 34} & 76 {\scriptsize $\pm$ 10} & 54 {\scriptsize $\pm$ 21} \\
& MedGemma & 70 {\scriptsize $\pm$ 7}  & 30 {\scriptsize $\pm$ 7}  & 0 {\scriptsize $\pm$ 0}  & 59 {\scriptsize $\pm$ 19} & 59 {\scriptsize $\pm$ 19} \\
& LLaVA    & 56 {\scriptsize $\pm$ 6}  & 44 {\scriptsize $\pm$ 6}  & 0 {\scriptsize $\pm$ 0}  & 35 {\scriptsize $\pm$ 28} & 35 {\scriptsize $\pm$ 28} \\
& LLaVA-Med & 50 {\scriptsize $\pm$ 0}  & 50 {\scriptsize $\pm$ 0}  & 0 {\scriptsize $\pm$ 0}  & 44 {\scriptsize $\pm$ 34} & 44 {\scriptsize $\pm$ 34} \\
\midrule
\multirow{4}{*}{Skin Cancer}
& Gemma    & 32 {\scriptsize $\pm$ 16} & 32 {\scriptsize $\pm$ 21} & 37 {\scriptsize $\pm$ 37} & 71 {\scriptsize $\pm$ 10} & 42 {\scriptsize $\pm$ 22} \\
& MedGemma & 58 {\scriptsize $\pm$ 7}  & 42 {\scriptsize $\pm$ 7}  & 0 {\scriptsize $\pm$ 0}  & 60 {\scriptsize $\pm$ 12} & 60 {\scriptsize $\pm$ 12} \\
& LLaVA    & 50 {\scriptsize $\pm$ 1}  & 50 {\scriptsize $\pm$ 1}  & 0 {\scriptsize $\pm$ 0}  & 36 {\scriptsize $\pm$ 33} & 36 {\scriptsize $\pm$ 33} \\
& LLaVA-Med & 49 {\scriptsize $\pm$ 4}  & 51 {\scriptsize $\pm$ 4}  & 0 {\scriptsize $\pm$ 0}  & 38 {\scriptsize $\pm$ 31} & 38 {\scriptsize $\pm$ 31} \\
\midrule
\multirow{4}{*}{IDC}
& Gemma    & 41 {\scriptsize $\pm$ 18} & 38 {\scriptsize $\pm$ 16} & 21 {\scriptsize $\pm$ 35} & 62 {\scriptsize $\pm$ 7}  & 51 {\scriptsize $\pm$ 24} \\
& MedGemma & 51 {\scriptsize $\pm$ 4}  & 49 {\scriptsize $\pm$ 4}  & 0 {\scriptsize $\pm$ 0}  & 38 {\scriptsize $\pm$ 26} & 38 {\scriptsize $\pm$ 26} \\
& LLaVA    & 51 {\scriptsize $\pm$ 4}  & 49 {\scriptsize $\pm$ 4}  & 0 {\scriptsize $\pm$ 0}  & 13 {\scriptsize $\pm$ 28} & 13 {\scriptsize $\pm$ 28} \\
& LLaVA-Med & 51 {\scriptsize $\pm$ 5}  & 49 {\scriptsize $\pm$ 5}  & 0 {\scriptsize $\pm$ 0}  & 23 {\scriptsize $\pm$ 32} & 23 {\scriptsize $\pm$ 32} \\
\bottomrule
\end{tabular}
\caption{Average performance across prompts for the \textbf{with answer in one word} setting. Values are reported as mean with standard deviation in smaller font. Corr = correct, Inc = incorrect, Ref = refused.}
\label{tab:all-diseases-oneword}
\end{table*}

\subsection{Prompt Variations Across Datasets}\label{app:prompts_var}

\begin{table}[t]
\centering
\small
\setlength{\tabcolsep}{4pt}
\begin{tabular}{llccccc}
\toprule
Model & Prompt Variant & Corr $\uparrow$ & Inc $\downarrow$ & Ref $\downarrow$ & F1$_{nr}$ $\uparrow$ & F1$_{all}$ $\uparrow$ \\
\midrule
Gemma 
 & A) Tumor \; B) No tumor & 58 & 42 & 0 & 70 & 70 \\
  
 & A) No tumor \; B) Tumor & 51 & 49 & 0 & 67 & 67 \\
  
 & Is the image normal? & 52 & 48 & 0 & 67 & 67 \\
  
 & Is the image abnormal? & 51 & 47 & 2 & 68 & 66 \\
  
 & Tumor present? (consider ...) & 59 & 27 & 14 & 78 & 67 \\
  
 & Tumor present? & 62 & 21 & 17 & 82 & 68 \\
\midrule
MedGemma 
 & \textbf{A) Tumor \; B) No tumor} & \textbf{91} & \textbf{9} & \textbf{0} & \textbf{91} & \textbf{91} \\
  
 & A) No tumor \; B) Tumor & 82 & 18 & 0 & 85 & 85 \\
  
 & Is the image normal? & 74 & 21 & 5 & 83 & 79 \\
  
 & Is the image abnormal? & 68 & 22 & 10 & 82 & 74 \\
  
 & Tumor present? (consider ...) & 72 & 14 & 14 & 88 & 75 \\
  
 & Tumor present? & 71 & 11 & 18 & 90 & 74 \\
\midrule
LLaVA 
 & A) Tumor \; B) No tumor & 68 & 20 & 12 & 81 & 71 \\
  
 & A) No tumor \; B) Tumor & 55 & 20 & 25 & 83 & 62 \\
  
 & Is the image normal? & 8 & 3 & 89 & 67 & 7 \\
  
 & Is the image abnormal? & 8 & 0 & 92 & 100 & 8 \\
  
 & Tumor present? (consider ...) & 6 & 0 & 94 & 100 & 6 \\
  
 & Tumor present? & 13 & 0 & 87 & 100 & 13 \\
\midrule
LLaVA-Med 
 & A) Tumor \; B) No tumor & 75 & 25 & 0 & 67 & 67 \\
  
 & A) No tumor \; B) Tumor & 65 & 35 & 0 & 74 & 74 \\
  
 & Is the image normal? & 64 & 36 & 0 & 44 & 44 \\
  
 & Is the image abnormal? & 54 & 46 & 0 & 15 & 15 \\
  
 & Tumor present? (consider ...) & 50 & 50 & 0 & 0 & 0 \\
  
 & Tumor present? & 50 & 42 & 8 & 0 & 0 \\
\bottomrule
\end{tabular}
\caption{Evaluation results for Brain Tumor without a one-word answer requirement. Corr = correct, Inc = incorrect, Ref = refused. $\uparrow$ indicates higher is better; $\downarrow$ indicates lower is better.}
\label{tab:brain-tumor-nooneword}
\end{table}

\begin{table}[t]
\centering
\small
\setlength{\tabcolsep}{4pt}
\begin{tabular}{llccccc}
\toprule
Model & Prompt Variant & Corr $\uparrow$ & Inc $\downarrow$ & Ref $\downarrow$ & F1$_{nr}$ $\uparrow$ & F1$_{all}$ $\uparrow$ \\
\midrule
Gemma 
 & A) Tumor \; B) No tumor & 62 & 35 & 3 & 74 & 72 \\
  
 & A) No tumor \; B) Tumor & 50 & 48 & 2 & 68 & 66 \\
  
 & Is the image normal? & 55 & 45 & 0 & 69 & 69 \\
  
 & Is the image abnormal? & 54 & 43 & 3 & 70 & 68 \\
  
 & Tumor present? (consider ...) & 71 & 23 & 6 & 81 & 76 \\
  
 & Tumor present? & 67 & 14 & 19 & 87 & 70 \\
\midrule
MedGemma 
 & \textbf{A) Tumor \; B) No tumor} & \textbf{90} & \textbf{10} & \textbf{0} & \textbf{90} & \textbf{90} \\
  
 & A) No tumor \; B) Tumor & 78 & 22 & 0 & 82 & 82 \\
  
 & Is the image normal? & 72 & 28 & 0 & 78 & 78 \\
  
 & Is the image abnormal? & 72 & 28 & 0 & 78 & 78 \\
  
 & Tumor present? (consider ...) & 80 & 20 & 0 & 83 & 83 \\
  
 & Tumor present? & 81 & 19 & 0 & 84 & 84 \\
\midrule
LLaVA 
 & A) Tumor \; B) No tumor & 60 & 40 & 0 & 33 & 33 \\
  
 & A) No tumor \; B) Tumor & 85 & 15 & 0 & 86 & 86 \\
  
 & Is the image normal? & 50 & 50 & 0 & 67 & 67 \\
  
 & Is the image abnormal? & 72 & 28 & 0 & 77 & 77 \\
  
 & Tumor present? (consider ...) & 77 & 23 & 0 & 71 & 71 \\
  
 & Tumor present? & 72 & 28 & 0 & 64 & 64 \\
\midrule
LLaVA-Med 
 & A) Tumor \; B) No tumor & 51 & 49 & 0 & 4 & 4 \\
  
 & A) No tumor \; B) Tumor & 77 & 23 & 0 & 80 & 80 \\
  
 & Is the image normal? & 60 & 40 & 0 & 33 & 33 \\
  
 & Is the image abnormal? & 60 & 40 & 0 & 33 & 33 \\
  
 & Tumor present? (consider ...) & 50 & 50 & 0 & 0 & 0 \\
  
 & Tumor present? & 50 & 50 & 0 & 0 & 0 \\
\bottomrule
\end{tabular}
\caption{Evaluation results for Brain Tumor with a one-word answer requirement. Corr = correct, Inc = incorrect, Ref = refused. $\uparrow$ indicates higher is better; $\downarrow$ indicates lower is better.}
\label{tab:brain-tumor-oneword}
\end{table}

\begin{table}[t]
\centering
\small
\setlength{\tabcolsep}{4pt}
\begin{tabular}{llccccc}
\toprule
Model & Prompt Variant & Corr $\uparrow$ & Inc $\downarrow$ & Ref $\downarrow$ & F1$_{nr}$ $\uparrow$ & F1$_{all}$ $\uparrow$ \\
\midrule
Gemma 
 & A) Pneumonia \; B) No Pneumonia & 51 & 49 & 0 & 67 & 67 \\
  
 & A) No Pneumonia \; B) Pneumonia & 51 & 49 & 0 & 67 & 67 \\
  
 & Is the image normal? & 55 & 45 & 0 & 69 & 69 \\
  
 & Is the image abnormal? & 52 & 48 & 0 & 68 & 68 \\
  
 & Pneumonia present? (consider ...) & 52 & 47 & 1 & 68 & 67 \\
  
 & Pneumonia present? & 50 & 50 & 0 & 67 & 67 \\
\midrule
MedGemma 
 & \textbf{A) Pneumonia \; B) No Pneumonia} & \textbf{63} & \textbf{37} & \textbf{0} & \textbf{41} & \textbf{41} \\
  
 & A) No Pneumonia \; B) Pneumonia & 62 & 38 & 0 & 39 & 39 \\
  
 & Is the image normal? & 49 & 12 & 39 & 68 & 42 \\
  
 & Is the image abnormal? & 35 & 43 & 22 & 62 & 48 \\
  
 & Pneumonia present? (consider ...) & 50 & 50 & 0 & 64 & 64 \\
  
 & Pneumonia present? & 0 & 0 & 100 & 0 & 0 \\
\midrule
LLaVA 
 & A) Pneumonia \; B) No Pneumonia & 50 & 45 & 5 & 0 & 0 \\
  
 & A) No Pneumonia \; B) Pneumonia & 50 & 45 & 5 & 0 & 0 \\
  
 & Is the image normal? & 0 & 0 & 100 & 0 & 0 \\
  
 & Is the image abnormal? & 0 & 0 & 100 & 0 & 0 \\
  
 & Pneumonia present? (consider ...) & 0 & 0 & 100 & 0 & 0 \\
  
 & Pneumonia present? & 0 & 0 & 100 & 0 & 0 \\
\midrule
LLaVA-Med 
 & A) Pneumonia \; B) No Pneumonia & 50 & 50 & 0 & 0 & 0 \\
  
 & A) No Pneumonia \; B) Pneumonia & 50 & 50 & 0 & 0 & 0 \\
  
 & Is the image normal? & 50 & 50 & 0 & 67 & 67 \\
  
 & Is the image abnormal? & 52 & 48 & 0 & 14 & 14 \\
  
 & Pneumonia present? (consider ...) & 50 & 50 & 0 & 67 & 67 \\
  
 & Pneumonia present? & 52 & 48 & 0 & 68 & 68 \\
\bottomrule
\end{tabular}
\caption{Evaluation results for Pneumonia without a one-word answer requirement. Corr = correct, Inc = incorrect, Ref = refused. $\uparrow$ indicates higher is better; $\downarrow$ indicates lower is better.}
\label{tab:pneumonia-nooneword}
\end{table}

\begin{table}[t]
\centering
\small
\setlength{\tabcolsep}{4pt}
\begin{tabular}{llccccc}
\toprule
Model & Prompt Variant & Corr $\uparrow$ & Inc $\downarrow$ & Ref $\downarrow$ & F1$_{nr}$ $\uparrow$ & F1$_{all}$ $\uparrow$ \\
\midrule
Gemma 
 & \textbf{A) Carcinoma \; B) No carcinoma} & \textbf{58} & \textbf{42} & \textbf{0} & \textbf{68} & \textbf{68} \\
  
 & A) No carcinoma \; B) Carcinoma & 53 & 47 & 0 & 68 & 68 \\
  
 & Is the image normal? & 26 & 15 & 59 & 62 & 25 \\
  
 & Is the image abnormal? & 27 & 29 & 44 & 62 & 35 \\
  
 & IDC present? (consider ...) & 34 & 26 & 40 & 70 & 42 \\
  
 & IDC present? & 13 & 17 & 70 & 59 & 18 \\
\midrule
MedGemma 
 & A) Carcinoma \; B) No carcinoma & 52 & 48 & 0 & 8 & 8 \\
  
 & A) No carcinoma \; B) Carcinoma & 50 & 50 & 0 & 51 & 51 \\
  
 & Is the image normal? & 15 & 13 & 72 & 32 & 9 \\
  
 & Is the image abnormal? & 21 & 14 & 65 & 71 & 25 \\
  
 & IDC present? (consider ...) & 2 & 0 & 98 & 100 & 2 \\
  
 & IDC present? & 3 & 0 & 97 & 100 & 3 \\
\midrule
LLaVA 
 & A) Carcinoma \; B) No carcinoma & 15 & 22 & 63 & 8 & 3 \\
  
 & A) No carcinoma \; B) Carcinoma & 9 & 5 & 86 & 44 & 6 \\
  
 & Is the image normal? & 1 & 2 & 97 & 0 & 0 \\
  
 & Is the image abnormal? & 1 & 3 & 96 & 0 & 0 \\
  
 & IDC present? (consider ...) & 19 & 15 & 66 & 72 & 24 \\
  
 & IDC present? & 23 & 30 & 47 & 61 & 32 \\
\midrule
LLaVA-Med 
 & A) Carcinoma \; B) No carcinoma & 0 & 0 & 100 & 0 & 0 \\
  
 & A) No carcinoma \; B) Carcinoma & 0 & 0 & 100 & 0 & 0 \\
  
 & Is the image normal? & 57 & 38 & 5 & 63 & 60 \\
  
 & Is the image abnormal? & 46 & 50 & 4 & 4 & 4 \\
  
 & IDC present? (consider ...) & 31 & 33 & 36 & 6 & 4 \\
  
 & IDC present? & 47 & 47 & 6 & 4 & 4 \\
\bottomrule
\end{tabular}
\caption{Evaluation results for IDC without a one-word answer requirement. Corr = correct, Inc = incorrect, Ref = refused. $\uparrow$ indicates higher is better; $\downarrow$ indicates lower is better.}
\label{tab:idc-nooneword}
\end{table}

\begin{table}[t]
\centering
\small
\setlength{\tabcolsep}{4pt}
\begin{tabular}{llccccc}
\toprule
Model & Prompt Variant & Corr $\uparrow$ & Inc $\downarrow$ & Ref $\downarrow$ & F1$_{nr}$ $\uparrow$ & F1$_{all}$ $\uparrow$ \\
\midrule
Gemma 
 & A) Carcinoma \; B) No carcinoma & 52 & 47 & 1 & 64 & 63 \\
  
 & A) No carcinoma \; B) Carcinoma & 52 & 48 & 0 & 68 & 68 \\
  
 & Is the image normal? & 52 & 48 & 0 & 65 & 65 \\
  
 & Is the image abnormal? & 54 & 46 & 0 & 67 & 67 \\
  
 & IDC present? (consider ...) & 29 & 29 & 42 & 61 & 36 \\
  
 & IDC present? & 9 & 8 & 83 & 50 & 8 \\
\midrule
MedGemma 
 & A) Carcinoma \; B) No carcinoma & 50 & 50 & 0 & 0 & 0 \\
  
 & A) No carcinoma \; B) Carcinoma & 52 & 48 & 0 & 40 & 40 \\
  
 & Is the image normal? & 56 & 44 & 0 & 66 & 66 \\
  
 & Is the image abnormal? & 54 & 46 & 0 & 68 & 68 \\
  
 & IDC present? (consider ...) & 46 & 54 & 0 & 21 & 21 \\
  
 & IDC present? & 47 & 53 & 0 & 33 & 33 \\
\midrule
LLaVA 
 & A) Carcinoma \; B) No carcinoma & 50 & 50 & 0 & 0 & 0 \\
  
 & A) No carcinoma \; B) Carcinoma & 59 & 41 & 0 & 70 & 70 \\
  
 & Is the image normal? & 49 & 51 & 0 & 4 & 4 \\
  
 & Is the image abnormal? & 50 & 50 & 0 & 0 & 0 \\
  
 & IDC present? (consider ...) & 50 & 50 & 0 & 0 & 0 \\
  
 & IDC present? & 48 & 52 & 0 & 4 & 4 \\
\midrule
LLaVA-Med 
 & A) Carcinoma \; B) No carcinoma & 46 & 54 & 0 & 7 & 7 \\
  
 & A) No carcinoma \; B) Carcinoma & 47 & 53 & 0 & 4 & 4 \\
  
 & \textbf{Is the image normal?} & \textbf{61} & \textbf{39} & \textbf{0} & \textbf{61} & \textbf{61} \\
  
 & Is the image abnormal? & 50 & 50 & 0 & 66 & 66 \\
  
 & IDC present? (consider ...) & 50 & 50 & 0 & 0 & 0 \\
  
 & IDC present? & 50 & 50 & 0 & 0 & 0 \\
\bottomrule
\end{tabular}
\caption{Evaluation results for IDC with a one-word answer requirement. Corr = correct, Inc = incorrect, Ref = refused. $\uparrow$ indicates higher is better; $\downarrow$ indicates lower is better.}
\label{tab:idc-oneword}
\end{table}

\begin{table}[t]
\centering
\small
\setlength{\tabcolsep}{4pt}
\begin{tabular}{llccccc}
\toprule
Model & Prompt Variant & Corr $\uparrow$ & Inc $\downarrow$ & Ref $\downarrow$ & F1$_{nr}$ $\uparrow$ & F1$_{all}$ $\uparrow$ \\
\midrule
Gemma 
 & A) Benign \; B) Malignant & 49 & 51 & 0 & 58 & 58 \\
  
 & A) Malignant \; B) Benign & 46 & 54 & 0 & 61 & 61 \\
  
 & Is the image normal? & 38 & 35 & 27 & 67 & 49 \\
  
 & Is the image abnormal? & 39 & 36 & 25 & 68 & 51 \\
  
 & Malignant skin cancer present? (consider ...) & 50 & 49 & 1 & 67 & 66 \\
  
 & Malignant skin cancer present? & 43 & 45 & 12 & 66 & 58 \\
\midrule
MedGemma 
 & \textbf{A) Benign \; B) Malignant} & \textbf{55} & \textbf{24} & \textbf{21} & \textbf{75} & \textbf{59} \\
  
 & A) Malignant \; B) Benign & 43 & 27 & 30 & 53 & 37 \\
  
 & Is the image normal? & 13 & 1 & 86 & 96 & 13 \\
  
 & Is the image abnormal? & 31 & 18 & 51 & 78 & 38 \\
  
 & Malignant skin cancer present? (consider ...) & 38 & 22 & 40 & 78 & 47 \\
  
 & Malignant skin cancer present? & 10 & 3 & 87 & 87 & 11 \\
\midrule
LLaVA 
 & A) Benign \; B) Malignant & 3 & 7 & 90 & 46 & 5 \\
  
 & A) Malignant \; B) Benign & 2 & 2 & 96 & 67 & 3 \\
  
 & Is the image normal? & 12 & 5 & 83 & 0 & 0 \\
  
 & Is the image abnormal? & 11 & 7 & 82 & 0 & 0 \\
  
 & Malignant skin cancer present? (consider ...) & 34 & 34 & 32 & 65 & 44 \\
  
 & Malignant skin cancer present? & 8 & 10 & 82 & 62 & 11 \\
\midrule
LLaVA-Med 
 & A) Benign \; B) Malignant & 44 & 56 & 0 & 30 & 30 \\
  
 & A) Malignant \; B) Benign & 50 & 50 & 0 & 47 & 47 \\
  
 & Is the image normal? & 50 & 50 & 0 & 66 & 66 \\
  
 & Is the image abnormal? & 50 & 50 & 0 & 67 & 67 \\
  
 & Malignant skin cancer present? (consider ...) & 41 & 55 & 4 & 57 & 55 \\
  
 & Malignant skin cancer present? & 50 & 49 & 1 & 0 & 0 \\
\bottomrule
\end{tabular}
\caption{Evaluation results for Skin Cancer without a one-word answer requirement. Corr = correct, Inc = incorrect, Ref = refused. $\uparrow$ indicates higher is better; $\downarrow$ indicates lower is better.}
\label{tab:skin-cancer-nooneword}
\end{table}

\begin{table}[t]
\centering
\small
\setlength{\tabcolsep}{4pt}
\begin{tabular}{llccccc}
\toprule
Model & Prompt Variant & Corr $\uparrow$ & Inc $\downarrow$ & Ref $\downarrow$ & F1$_{nr}$ $\uparrow$ & F1$_{all}$ $\uparrow$ \\
\midrule
Gemma 
 & A) Benign \; B) Malignant & 46 & 54 & 0 & 63 & 63 \\
  
 & A) Malignant \; B) Benign & 45 & 55 & 0 & 61 & 61 \\
  
 & Is the image normal? & 39 & 37 & 24 & 68 & 52 \\
  
 & Is the image abnormal? & 28 & 18 & 54 & 76 & 35 \\
  
 & Malignant skin cancer present? (consider ...) & 28 & 25 & 47 & 69 & 37 \\
  
 & Malignant skin cancer present? & 4 & 1 & 95 & 89 & 4 \\
\midrule
MedGemma 
 & A) Benign \; B) Malignant & 62 & 38 & 0 & 50 & 50 \\
  
 & A) Malignant \; B) Benign & 61 & 39 & 0 & 40 & 40 \\
  
 & Is the image normal? & 50 & 50 & 0 & 67 & 67 \\
  
 & Is the image abnormal? & 49 & 51 & 0 & 66 & 66 \\
  
 & \textbf{Malignant skin cancer present? (consider ...)} & \textbf{64} & \textbf{35} & \textbf{1} & \textbf{69} & \textbf{68} \\
  
 & Malignant skin cancer present? & 63 & 37 & 0 & 68 & 68 \\
\midrule
LLaVA 
 & A) Benign \; B) Malignant & 49 & 51 & 0 & 66 & 66 \\
  
 & A) Malignant \; B) Benign & 51 & 49 & 0 & 4 & 4 \\
  
 & Is the image normal? & 50 & 50 & 0 & 67 & 67 \\
  
 & Is the image abnormal? & 50 & 50 & 0 & 67 & 67 \\
  
 & Malignant skin cancer present? (consider ...) & 51 & 49 & 0 & 8 & 8 \\
  
 & Malignant skin cancer present? & 50 & 50 & 0 & 7 & 7 \\
\midrule
LLaVA-Med 
 & A) Benign \; B) Malignant & 41 & 59 & 0 & 37 & 37 \\
  
 & A) Malignant \; B) Benign & 53 & 47 & 0 & 61 & 61 \\
  
 & Is the image normal? & 48 & 51 & 1 & 64 & 64 \\
  
 & Is the image abnormal? & 50 & 50 & 0 & 67 & 67 \\
  
 & Malignant skin cancer present? (consider ...) & 50 & 50 & 0 & 0 & 0 \\
  
 & Malignant skin cancer present? & 50 & 50 & 0 & 0 & 0 \\
\bottomrule
\end{tabular}
\caption{Evaluation results for Skin Cancer with a one-word answer requirement. Corr = correct, Inc = incorrect, Ref = refused. $\uparrow$ indicates higher is better; $\downarrow$ indicates lower is better.}
\label{tab:skin-cancer-oneword}
\end{table}
\end{document}